\documentclass[letterpaper]{article} 
\usepackage{aaai25}  
\usepackage{times}  
\usepackage{helvet}  
\usepackage{courier}  
\usepackage[hyphens]{url}  
\usepackage{graphicx} 
\usepackage{natbib}  
\usepackage{caption} 
\usepackage{algorithm}
\usepackage{algorithmic}
\usepackage{amsthm}
\usepackage{url}            
\usepackage{amsfonts}       
\usepackage{nicefrac}       
\usepackage{microtype}      
\usepackage{xcolor}         
\usepackage{amsmath}
\usepackage{ragged2e} 
\usepackage{booktabs,makecell, multirow, tabularx}
\usepackage{enumitem}
\usepackage{amsthm}
\usepackage{mathtools}
\usepackage{subfigure}
\usepackage{footmisc}
\usepackage{colortbl}
\usepackage{nicematrix}
\usepackage{caption}
\usepackage{array}
\usepackage{subcaption}

\usepackage{newfloat}
\usepackage{listings}

\DeclareCaptionStyle{ruled}{labelfont=normalfont,labelsep=colon,strut=off} 
\floatstyle{ruled}
\newfloat{listing}{tb}{lst}{}
\floatname{listing}{Listing}
\title{Diffusion Prior Interpolation for Flexibility Real-World Face Super-Resolution}

\author{
    Jiarui Yang\textsuperscript{\rm 1, \rm 2},
    Tao Dai\textsuperscript{\rm 3, \thanks{Corresponding author: daitao.edu@gmail.com.}},
    Yufei Zhu\textsuperscript{\rm 3},
    Naiqi Li\textsuperscript{\rm 2},
	Jinmin Li\textsuperscript{\rm 2},
    Shu-Tao Xia\textsuperscript{\rm 2}
}
\affiliations{
    \textsuperscript{\rm 1}College of Artificial Intelligence, Nankai University, Tianjin, China \\
    \textsuperscript{\rm 2}Tsinghua Shenzhen International Graduate School, Tsinghua University, Shenzhen, China \\
    \textsuperscript{\rm 3}College of Computer Science and Software Engineering, Shenzhen University, China \\
    1120230264@mail.nankai.edu.cn, daitao.edu@gmail.com \\
    zhuyufei2021@email.szu.edu.cn, ljm22@mails.tsinghua.edu.cn, \{linaiqi, xiast\}@sz.tsinghua.edu.cn
}

\usepackage{bibentry}

\begin{document}

\maketitle

\begin{abstract}
Diffusion models represent the state-of-the-art in generative modeling. Due to their high training costs, many works leverage pre-trained diffusion models' powerful representations for downstream tasks, such as face super-resolution (FSR), through fine-tuning or prior-based methods. However, relying solely on priors without supervised training makes it challenging to meet the pixel-level accuracy requirements of discrimination task. Although prior-based methods can achieve high fidelity and high-quality results, ensuring consistency remains a significant challenge. In this paper, we propose a masking strategy with strong and weak constraints and iterative refinement for real-world FSR, termed Diffusion Prior Interpolation (DPI). We introduce conditions and constraints on consistency by masking different sampling stages based on the structural characteristics of the face. Furthermore, we propose a condition Corrector (CRT) to establish a reciprocal posterior sampling process, enhancing FSR performance by mutual refinement of conditions and samples. DPI can balance consistency and diversity and can be seamlessly integrated into pre-trained models. In extensive experiments conducted on synthetic and real datasets, along with consistency validation in face recognition, DPI demonstrates superiority over SOTA FSR methods. The code is available at \url{https://github.com/JerryYann/DPI}.
\end{abstract}

%

\section{Introduction}
Image super-resolution (SR) is a classic ill-posed problem aimed at enhancing image quality by restoring high-resolution (HR) details from low-resolution (LR) images. In the context of face super-resolution (FSR), this technology is applied in areas such as face recognition~\cite{zou2011very} and visual enhancement~\cite{jiang2021deep}. These applications typically require SR images to exhibit both high consistency and fidelity. Previous SR work has tended to overly pursue distortion-based quantitative metrics (such as PSNR, SSIM)~\cite{chen2018fsrnet,gao2023ctcnet}. Blau et al.~\cite{blau2018perception} mathematically prove that distortion and perceptual quality are at odds with each other, implying that excessive pursuit of PSNR or SSIM indirectly leads to poorer fidelity. Moreover, discriminative model-based SR methods~\cite{gao2023ctcnet, wang2023spatial} typically employ end-to-end training, achieving it by minimizing pixel-wise loss between the SR output image and the GT image. It is well-known that such learning objectives favor distortion measures and constrain the SR output to the average of multiple possibilities. This maintains a certain consistency but potentially results in over-smoothing outputs~\cite{sajjadi2017enhancenet}, as shown in Fig. \ref{fig1}(c-d).

In contrast, generative methods such as Variational Autoencoders (VAEs)~\cite{liu2021variational}, Normalizing Flows (NFs)~\cite{lugmayr2020srflow}, Generative Adversarial Networks (GANs)~\cite{wang2023gan} and Diffusion Models (DMs)~\cite{ho2020denoising, song2020denoising} have the capability to generate high-fidelity images. Among these, Denoising Diffusion Probabilistic Models (DDPMs)~\cite{ho2020denoising} have recently gained significant attention and research interest due to their impressive generative capabilities. DDPMs exhibit advantages such as stable training and enhanced controllability compared to other generative models~\cite{brock2018large, kirichenko2020normalizing, bredell2023explicitly}. At present, diffusion-based FSR work can be broadly categorized into those that require task-directed retraining~\cite{saharia2023image, wei2023raindiffusion, li2022srdiff,whang2022deblurring} and prior-based methods~\cite{choi2021ilvr, chung2022diffusion, kawar2022denoising}. Training a conditional DDPM from scratch requires significant computational resources and can limit the prior space to lead to sub-optimal results~\cite{StableSR}. Introducing conditions to utilize the priors encapsulated in the pre-trained model is an alternating solution. However, adding conditions to a pre-trained model introduces errors and visual artifacts in the intrinsic probability distributions at each time step, leading to the generation of results that deviate from the model's prior manifold~\cite{mei2022bi}. As shown in Fig. \ref{fig1}(e-f), although these prior-based methods achieve good visual perception, the issue of consistency remains unresolved. This is primarily because prior-based methods are unsupervised and generative in nature. When dealing with low-level tasks requiring pixel-level accuracy, it is challenging to achieve precise discrimination.

\begin{figure*}[ht]
    \centering
    \includegraphics[height=0.350\textwidth]{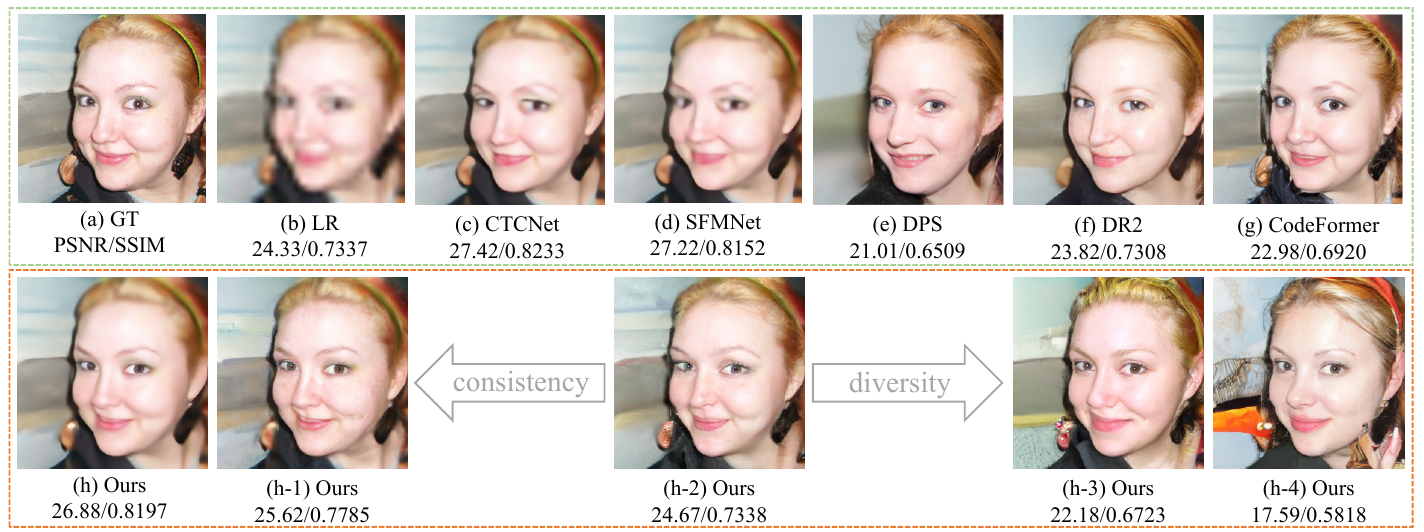}
    \caption{Visualization of FSR results using different types of methods, including those based on discriminative models (c-d)~\cite{gao2023ctcnet, wang2023spatial}, generative models (e-f)~\cite{chung2022diffusion, wang2023dr2}, and prior-based methods (e-h)~\cite{zhou2022towards}. Our approach combines the strengths of discriminative and generative models, allowing for flexible adjustment of facial diversity and fidelity.} 
    \label{fig1}
    \vspace{-0.51cm}
\end{figure*}

We are aware that the sampling process of DDPMs is an iterative one, progressing from coarse to fine. Wang et al.~\cite{wang2023dr2} have been demonstrated that there exists a time step that partitions the sampling interval, and beyond this time step, the error between the real posterior distribution and the posterior distribution introduced by conditioning becomes sufficiently small. This means that during the early sampling phase, the impact of the conditions is more significant because the distribution of conditions is closer to the true distribution. This inspire us to impose strong and weak constraints on the posterior distribution to flexibly regulate the consistency and diversity of FSR, as depicted in the lower part of Fig. \ref{fig1}. We maintain the condition distribution to match the real posterior distribution, alleviating cumulative errors in the distribution and deviations from the prior manifold. In addition, we combine the advantages of discriminative methods to construct an accurate and refined sampling process. 

Specifically, we take the LR image as a condition and design various masks tailored to facial features, combining them to form different Condition Masks (CMs). The CMs include a Fixed Condition Mask (FCM) and a Randomly Adaptive Condition Mask (RACM), to mask the sampling process. This is similar to the prior-based inpainting methods~\cite{song2020score, lugmayr2022repaint}, but our CMs can provide detailed neighborhood information to better utilize the priors. We use an adjustable scalar to divide the sampling process into two stages, where the first stage utilizes the FCM to limit the prior space and ensure consistency in FSR. In the second stage, the RACM, guided by facial structure supervision, ensures consistency while enhancing the fidelity and diversity of the sampling. The trade-off between consistency and fidelity can be realized by adjusting this scalar value. In addition, we introduce a condition Corrector (CRT) to create a reciprocal sampling process where conditions are updated with the assistance of priors, and samples improve with the refinement of conditions. The CRT trades a very small time cost for better performance. Our approach leverages the priors of diffusion to interpolate the condition masks, which we call Diffusion Prior Interpolation (DPI). \textbf{Our main contributions are summarized as follows}:
\begin{itemize}
    \item We propose DPI, which effectively leverages the priors of pre-trained diffusion models for real-world FSR. Extensive experiments on both synthetic and real-world datasets demonstrate that our method outperforms SOTA FSR methods. Furthermore, we validate the consistency of the SR images using a face recognition model, where our method also achieves the best results.
    \item We propose a novel masking strategy tailored for facial features to mask the diffusion sampling process. Our approach ensures structural consistency and detail diversity in the face while supporting flexible sampling.
    \item By introducing a condition CRT, we establish a reciprocal process between conditions and samples. CRT incurs a small time cost in exchange for improved performance. CRT is not limited to specific networks and endows DPI with scalability.
\end{itemize}

\begin{figure*}[ht]
    \centering
    \includegraphics[height=0.276\textwidth]{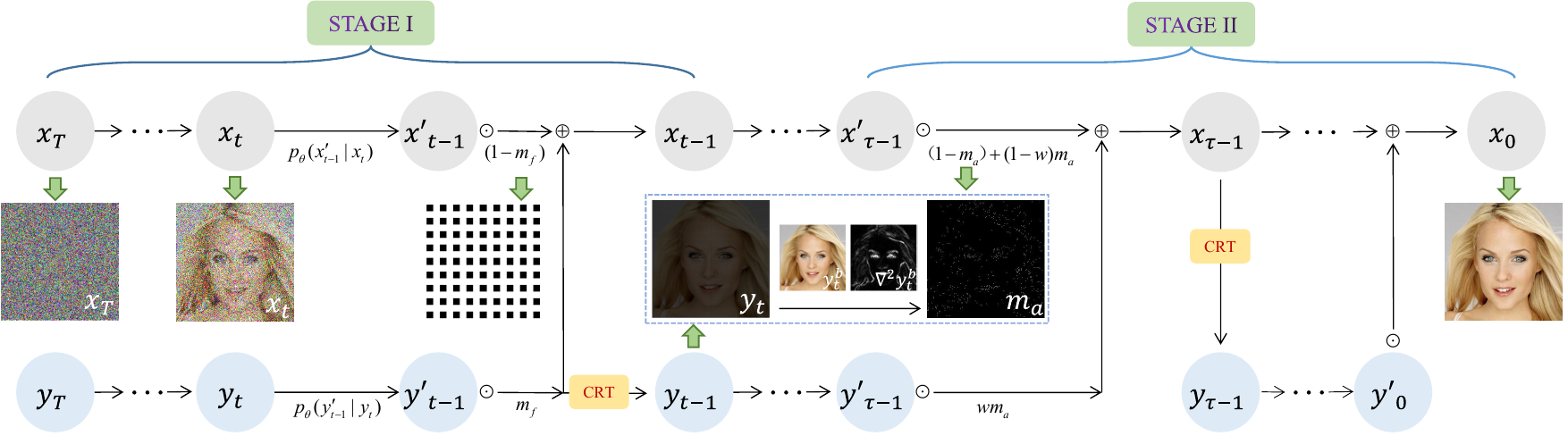}
    \caption{\textbf{Graphical model of Diffusion Prior Interpolation.} $\odot$ represents element-wise matrix multiplication. $\boldsymbol{x}_{T}$ and $\boldsymbol{y}_{T}$ correspond to the initial random noise and the initial condition respectively. We use the scalar $\tau$ to divide the sampling process into two stages. CRT is the Corrector function that applies Eq. \ref{eq:16} to CMs correction. $\boldsymbol{y}_{t}$ represents the intermediate condition. After posterior sampling, $\boldsymbol{y}'_{t}$ is multiplied by $\boldsymbol{m}_{f}$ and $\boldsymbol{m}_{a}$ to obtain CMs, including the FCM and RACM. Algorithm \ref{alg1} provides a detailed description of our DPI.} \vspace{-0.5cm}
    \label{fig2}
\end{figure*}

\section{Related work}
\subsection{Prior-based Face Super-Resolution}

Face images possess distinctive characteristics such as subject-centered focus, prominent foreground-background contrast, and well-defined face structures. Leveraging these prior information in previous work has effectively improved FSR performance~\cite{menon2020pulse,chen2021progressive,leng2022rcnet}. FSRCH~\cite{lu2022rethinking} introduces a pre-prior guiding method that extracts face priors from HR images and incorporates it into LR inputs, generating LRmix as a new SR input. Wang et al.~\cite{wang2022propagating} employ distillation to propagate real face priors learned by a teacher network to guide the learning of a student network for FSR. In order to enhance the ability of face restoration, Wang et al.~\cite{wang2022restoreformer} propose a Restoreformer that utilizes a high-quality dictionary that not only provides priors for the face, nose, and mouth but is generated through a high-quality face network that learns from a large number of undegraded faces. GLEAN~\cite{chan2022glean} directly leverages rich and diverse priors encapsulated in a pre-trained face GAN. Zhou et al.~\cite{zhou2022towards} introduce CodeFormer, a transformer-based prediction network that utilizes discrete codebooks learned in a compact proxy space through blind face recovery. The primary goal of CodeFormer is to reduce the uncertainty and ambiguity associated with recovery mapping by employing code prediction as a task. 

We design CMs based on the unique features of the face, such as facial contours. As a result, our masking strategy effectively ensures consistent sampling of facial structures. Additionally, by leveraging the priors from pre-trained models, DPI enhances the fidelity and quality of FSR.

\subsection{Conditional Diffusion Models}

In the spatial domain, concatenating LR image to train a DDPM from scratch is a simple and efficient SR method, such as SR3~\cite{saharia2023image} and SRDiff~\cite{li2022srdiff}. Whang et al.~\cite{whang2022deblurring} train a DDPM to learn the residual between MSE-estimated images and HR images to enhance the diversity of the MSE-estimated images. RainDiffusion~\cite{wei2023raindiffusion} employs two DDPMs for generating clear and degraded image pairs and then utilizes an idea similar to SR3 to achieve real-world image restoration. However, these conditional methods necessitate retraining the DDPM, which can be costly. To mitigate this, Brian et al.~\cite{moser2023dwa} propose training a DDPM in the discrete wavelet domain to learn conditional residual information, effectively reducing model parameters and achieving good SR performance. Additionally, most works introduce conditions into pre-trained DDPMs to circumvent training from scratch. PGDiff~\cite{yang2024pgdiff} utilizes the structure and color statistics of reference images to alleviate severe degradation issues. ILVR~\cite{choi2021ilvr} incorporates low-frequency information from reference images into the posterior distribution to control the conditional generation of pre-trained DDPMs. Fei et al.~\cite{fei2023generative} propose generating diffusion priors to model the posterior distribution of a pre-trained DDPM through unsupervised sampling. DPS~\cite{chung2022diffusion} applies conditional gradient correction to the posterior for guiding the sampling process. Methods based on pre-trained models are influenced by conditional intensity and controllability; if the condition intensity or control is weak, the fidelity of the results is higher, but consistency is lacking, and vice versa.

\section{Method}
\subsection{Preliminary}
DDPMs define a Markovian forward diffusion process with $T$ steps, which continuously transforms the initial state $\boldsymbol{x}_0$ through pre-specified noise scheduler $\{\beta_1, \beta_2, ..., \beta_T\}$ into an isotropic Gaussian distribution $\boldsymbol{x}_T \sim \mathcal{N}(\mathbf{0}, \mathbf{I})$. Each step of the forward process can be represented as a Gaussian transition, denoted as: 
\begin{equation}
    q(\boldsymbol{x}_t|\boldsymbol{x}_{t-1})= \mathcal{N}(\boldsymbol{x}_{t}|\sqrt{\alpha_t} \boldsymbol{x}_{t-1},(1-\alpha_t)\mathbf{I})
    \label{eq:1}
\end{equation}
where $\alpha_t=1-\beta_t$. Additionally, we can obtain the state at step $t$ through a single-step transition:
\begin{equation}
    q(\boldsymbol{x}_{t}|\boldsymbol{x}_0)= \mathcal{N}(\boldsymbol{x}_{t}|\sqrt{\bar \alpha_t}\boldsymbol{x}_0,(1- \bar \alpha_t)\mathbf{I})
    \label{eq:2}
\end{equation}
where $\bar \alpha_t={\prod_{i=1}^{T}} \alpha_t$. The inverse process starts from $\mathcal{N}(\mathbf{0}, \mathbf{I})$ and employs a U-Net denoiser~\cite{ronneberger2015u} with learnable parameters $\theta$ to fit the true posterior distribution $q(\boldsymbol{x}_{t-1} | \boldsymbol{x}_{t})$. A judicious noise scheduler is employed to ensure that the inverse process also follows a Gaussian distribution~\cite{bartlett1978introduction}, which can be represented as:
\begin{equation}
    p_\theta(\boldsymbol{x}_{t-1}|\boldsymbol{x}_{t})= \mathcal{N}(\boldsymbol{x}_{t-1}|\boldsymbol{\mu}_{\theta}(\boldsymbol{x}_t,t), \boldsymbol{\Sigma}_{\theta}(\boldsymbol{x}_{t},t))
    \label{eq:3}
\end{equation}
where $\boldsymbol{\mu}_{\theta}(\boldsymbol{x}_t,t)$ and $\boldsymbol{\Sigma}_{\theta}(\boldsymbol{x}_t,t)$ are the mean and variance predicted by the denoiser, respectively. Ho et al.~\cite{ho2020denoising} observe that directly predicting the mean is not optimal. Instead, a prevalent methodology involves parameterizing the $\boldsymbol{\mu}_{\theta}(\boldsymbol{\hat x}_0,\boldsymbol{x}_t,t)$ using a simplified loss function $\mathcal{L}_{simple}=||\boldsymbol{\epsilon} - \boldsymbol{\epsilon}_{\theta}||^{2}_{2}$ to predict the noise $\boldsymbol{\epsilon}_\theta$. This can be represented by the following formula:
\begin{equation}
    \boldsymbol{\mu}_{\theta}(\boldsymbol{\hat x}_0,\boldsymbol{x}_t,t) = \frac{\sqrt{\bar \alpha_t}(1-\bar \alpha_{t-1})}{1-\bar \alpha_t}\boldsymbol{x}_t + \frac{\sqrt{\bar \alpha_{t-1}}\beta_t}{1 - \bar \alpha_{t}}\boldsymbol{\hat x}_0
    \label{eq:4}
\end{equation}
where $\boldsymbol{\hat x}_0 = \frac{1}{\sqrt{\bar \alpha_t}}(\boldsymbol{x}_t - \sqrt{1 - \bar \alpha_t}\boldsymbol{\epsilon}_\theta(\boldsymbol{x}_t, t))$. Furthermore, Dhariwal et al.~\cite{dhariwal2021diffusion} propose that incorporating an additional output $v$ from the denoiser to parameterize the variance $\boldsymbol{\Sigma}_{\theta}(\boldsymbol{x}_t,t)$ is superior to the conventional approach of using a fixed variance $\beta_t$, we get:
\begin{equation}
    \boldsymbol{\Sigma}_{\theta}(\boldsymbol{x}_t,t) = exp(vlog\beta_t + (1-v)log\tilde\beta_t)
    \label{eq:5}
\end{equation}
where $\tilde \beta_t = \frac{1-\bar \alpha_{t-1}}{1-\bar\alpha_t}\beta_t$. Both improved algorithms like IDDPM~\cite{nichol2021improved} and accelerated sampling algorithms like DDIM~\cite{song2020denoising} follow a similar paradigm. Our proposed DPI can be plug-and-play in this paradigm.

\begin{algorithm}[t]
\small
\caption{Diffusion Prior Interpolation, given a diffusion model $(\boldsymbol{\mu}_{\theta}(\cdot), \boldsymbol{\Sigma}_{\theta}(\cdot))$ and Corrector $CRT(\cdot)$.}
\label{alg1}
\textbf{Input}: $\boldsymbol{y}_T$, $\tau$, $s$, $\omega$, $\boldsymbol{m}_{f}$ \\
$\boldsymbol{x}_T \gets$ sample from $\mathcal{N}(\mathbf{0}, \mathbf{I})$
\begin{algorithmic}[1] 
\FORALL{$t$ from $T$ to 1}
\STATE $\boldsymbol{\mu}_{x_t}, \boldsymbol{\Sigma}_{x_t} \gets \boldsymbol{\mu}_{\theta}(\boldsymbol{\hat{x}}_0, \boldsymbol{x}_t, t), \boldsymbol{\Sigma}_{\theta}(\boldsymbol{x}_t, t)$
\STATE $\boldsymbol{x}'_{t-1} \gets$ sample from $\mathcal{N}(\boldsymbol{\mu}_{x_t}, \boldsymbol{\Sigma}_{x_t})$
\STATE $\boldsymbol{y}_t^{n} = \sqrt{\bar{\alpha}_t}\boldsymbol{y}_t + \sqrt{1-\bar{\alpha}_t}\boldsymbol{\epsilon}_{\theta}$
\STATE $\boldsymbol{\mu}_{y_t} = \boldsymbol{\mu}_\theta(\boldsymbol{y}_t, \boldsymbol{y}_t^n, t)$
\STATE $\boldsymbol{y}'_{t-1} \gets$ sample from $\mathcal{N}(\boldsymbol{\mu}_{y_t}, \boldsymbol{\Sigma}_{x_t})$
\IF{$t > \tau$}
\STATE $\boldsymbol{x}_{t-1} = (1 - \boldsymbol{m}_{f}) \odot \boldsymbol{x}'_{t-1} + \boldsymbol{m}_{f} \odot \boldsymbol{y}'_{t-1}$
\ELSE
\STATE $\boldsymbol{m}_{a} = Mask_{gen}(\boldsymbol{y}_{t}, s)$
\STATE $w = t / \omega$
\STATE 
$\boldsymbol{x}_{t-1} = (1- \boldsymbol{m}_{a}) \odot \boldsymbol{x}'_{t-1} + w  \boldsymbol{m}_{a} \odot \boldsymbol{y}'_{t-1} +$ \\
\quad \quad \quad$ (1-w) \boldsymbol{m}_{a} \odot \boldsymbol{x}'_{t-1}$
\STATE $\boldsymbol{y}'_{t-1} = \boldsymbol{x}_{t-1}$
\ENDIF
\STATE $\boldsymbol{y}_{t-1} = CRT(\boldsymbol{m}_{f} \odot \boldsymbol{y}'_{t-1}, \boldsymbol{y}_T, t)$
\ENDFOR
\RETURN $\boldsymbol{x}_0$
\end{algorithmic}
\end{algorithm}

\subsection{Diffusion Prior Interpolation}

As shown in the upper part of Fig. \ref{fig2}, it represents an unconditional diffusion sampling process. Conditions are introduced through masking, as illustrated in the lower half. The conditions consist of an initial condition $\boldsymbol{y}_T$ and intermediate conditions $\boldsymbol{y}_t$. We design a fixed mask $\boldsymbol{m}_f$ and an adaptive mask $\boldsymbol{m}_a$ for generating Condition Masks (CMs). We will describe the form of CMs in detail in the following subsection.

The prior knowledge of the DDPMs is encapsulated in the denoiser, requiring alignment of the condition with the noise of the posterior distribution to effectively leverage the priors. We start by forward sampling the condition, and we get:
\begin{equation}
    \boldsymbol{y}_t^n = \sqrt{\bar \alpha_t}\boldsymbol{y}_t + \sqrt{1 - \bar \alpha_t} \boldsymbol{\epsilon}_{\theta}(\boldsymbol{x}_t,t)
    \label{eq:10}
\end{equation}
where $\boldsymbol{y}_t^n$ represents the noisy condition obtained through the reparameterization of Eq. \ref{eq:2} and $\boldsymbol{\epsilon}_{\theta}$ corresponds to the noise predicted by the pre-trained DDPM. Subsequently, we sample the conditional posterior distribution $\boldsymbol{y}'_{t-1}$ through the following equation:
\begin{equation}
    p_\theta(\boldsymbol{y}'_{t-1}|\boldsymbol{y}_{t})= \mathcal{N}(\boldsymbol{x}_{t-1}|\boldsymbol{\mu}_{\theta}(\boldsymbol{y}_t,\boldsymbol{y}_t^n,t), \boldsymbol{\Sigma}_{\theta}(\boldsymbol{x}_{t},t))
    \label{eq:7}
\end{equation}
where the mean and variance are aligned with the posterior distribution of the pre-trained DDPM ($\boldsymbol{x}'_{t-1}$). We divide the sampling interval by a scalar $\tau$. In the first stage, i.e. for $t \geq \tau$, we mask the $\boldsymbol{x}'_{t-1}$ to constrain the sample space, ensuring consistent sampling. We obtain:
\begin{equation}
    \boldsymbol{x}_{t-1} = (1-\boldsymbol{m}_{f}) \odot \boldsymbol{x}'_{t-1} + \boldsymbol{m}_{f} \odot \boldsymbol{y}'_{t-1}
    \label{8}
\end{equation}
where $\boldsymbol{m}_{f} \odot \boldsymbol{y}'_{t-1}$ is the Fixed Condition Mask (FCM) and $\boldsymbol{x}_{t-1}$ denotes the new posterior distribution obtained from this CM masking. $\boldsymbol{x}'_{t-1}$ are injected with conditional information, and in the subsequent sampling, the denoising prior is utilized to interpolate the $\boldsymbol{x}_{t-1}$. In the second stage, fidelity is further enhanced by incorporating a weighted Randomly Adaptive Condition Mask (RACM). When $t < \tau$, the condition is added as follows:
\begin{equation}  
    \boldsymbol{x}_{t-1} = (1 - \boldsymbol{m}_{a}) \odot \boldsymbol{x}'_{t-1} + w \boldsymbol{m}_{a} \odot \boldsymbol{y}'_{t-1} + (1-w)\boldsymbol{m}_{a} \odot \boldsymbol{x}'_{t-1}
    \label{eq:9}
\end{equation}
where $w = t / \omega$ represents a time-dependent weight controlled by the parameter $\omega$ and $\boldsymbol{m}_{a} \odot \boldsymbol{y}'_{t-1}$ is the RACM. $w$ gradually decreases over time steps, leading to a reduction in the intensity of the conditioning.

\subsection{Condition Masks}
\label{sec3.2}
In this section, we will mainly discuss the forms of the CMs and how to design both the strongly constrained FCM and the weakly constrained RACM. With the constraints of different CMs, we can flexibly realize consistent sampling and diversity generation.

Given an LR image $\boldsymbol{I}_{L}$ of size ($h, w$), the objective of FSR is to upscale it to the size ($H, W$) of HR image $\boldsymbol{I}_{H}$ with a scale factor of $H/h$. To achieve this, we first upsample $\boldsymbol{I}_{L}$ using Bicubic interpolation to the size ($H/k, W/k$), which serves as the base condition $\boldsymbol{I}_{L}^{bc}$. Here, the parameter $k$ controls sparsity. Subsequently, we design different forms of masks for the two stages of posterior sampling. Specifically, in the first stage, a fixed mask $\boldsymbol{m}_{f}$ is designed as shown in Fig. \ref{fig2}. The specific definition of $\boldsymbol{m}_{f}$ is as follows:

\begin{equation}
\boldsymbol{m}_{f}(i, j)=
\left\{
\begin{aligned}
1, &~if~~i,j \bmod k=0 \\
0,&~otherwise\\
\end{aligned}
\right.
\label{eq:10}
\end{equation}
where $\boldsymbol{m}_{f} \in R^{H\times W}$. When $k$ is set to 2, the form of $\boldsymbol{m}_{f}$ resembles a grid mask~\cite{chen2020gridmask} with a grid size of 1 pixel. We project $\boldsymbol{I}_{L}^{bc}$ onto $\boldsymbol{m}_{f}$ to generate an initial condition, denoted as $\boldsymbol{y}_T$:
\begin{equation}
\boldsymbol{y}_{T}(i, j)=
\left\{
\begin{aligned}
\boldsymbol{I}_{L}^{bc}(\left\lfloor \frac{i}{k} \right\rfloor, \left\lfloor \frac{j}{k} \right\rfloor), &~if~~\boldsymbol{m}_{f}(i,j)=1 \\
0,&~otherwise\\
\end{aligned}
\right.
\label{eq:11}
\end{equation}
where $\boldsymbol{I}_{L}^{bc} \in R^{\frac{H}{k}\times \frac{W}{k}}$ and $\boldsymbol{y}_{T} \in R^{H\times W}$. The $\boldsymbol{y}_T$ is initially composed of information from the LR image, then is updated to an intermediate condition $\boldsymbol{y}_t$ by CRT during the sampling phase. In the second stage, we backtrack the intermediate condition $\boldsymbol{y}_t$ and we obtain $\boldsymbol{y}_{t}^b$:

\begin{equation}
\begin{aligned}    
    \boldsymbol{y}_{t}^b(i, j) = \boldsymbol{y}_{t}(ki,kj)
    \label{eq:12}
\end{aligned}
\end{equation}
where $\boldsymbol{y}_{t} \in R^{H\times W}$ and $\boldsymbol{y}_{t}^b \in R^{\frac{H}{k}\times \frac{W}{k}}$. In Fig. \ref{fig2}, we illustrate the visualization of $\boldsymbol{y}_{t}^b$. Then, we extract an edge map $\nabla^{2} \boldsymbol{y}_{t}^b$ using the first-order Laplacian operator, which exhibits characteristics where pixel values are smaller in low-frequency regions and larger in high-frequency regions. Next, we normalize the edge map to the range of (0-1) to obtain a probability map $\boldsymbol{p}$:
\begin{equation}
\begin{aligned}    
    \boldsymbol{p} = \frac{\nabla^{2} \boldsymbol{y}_{t}^b - \text{min}\nabla^{2} \boldsymbol{y}_{t}^b}{\text{max}\nabla^{2} \boldsymbol{y}_{t}^b - \text{min}\nabla^{2} \boldsymbol{y}_{t}^b}
    \label{eq:13}
\end{aligned}
\end{equation}

For each pixel value $\boldsymbol{p}(i, j)$, we use it as a probability and then project it to $\boldsymbol{m}_{f}$ to generate a randomly adaptive mask $\boldsymbol{m}_{a}$ as follows:
\begin{equation}
\begin{aligned}    
    \boldsymbol{m}_{a}(i,j) = \{&0, ~if~~ \boldsymbol{m}_{f}(i,j)=0,\\
               &1, otherwise ~ \text{with probability } \boldsymbol{p}(\left\lfloor \frac{i}{k} \right\rfloor, \left\lfloor \frac{j}{k} \right\rfloor)^s\}
    \label{eq:14}
\end{aligned}
\end{equation}
where $\boldsymbol{m}_{a} \in R^{H\times W}$ and $s$ adjusts the probability distribution. We refer to the aforementioned process as $Mask_{gen}(\boldsymbol{y}_t, s)$. It's important to note that a different $\boldsymbol{m}_{a}$ is generated at each time step. We generate the RACM by masking the conditional posterior distribution $\boldsymbol{y}'_{t-1}$ using $\boldsymbol{m}_{a}$. The RACM exhibits high sparsity and features similar to the edge guidance in ControlNet~\cite{zhang2023adding}. This amplifies the prior space, enabling the generation of more texture details, thereby enhancing the fidelity and diversity of the results.

\subsection{Condition Corrector}
In the real world, LR images often suffer from various unknown degradations, leading to conditions that significantly deviate from the prior manifold. We propose a condition Corrector (CRT), to pull back the conditions to the prior space and establish a reciprocal sampling, as illustrated in Fig. \ref{fig3}. CRT is a small neural network conditioned on the initial condition $\boldsymbol{y}_{T}$, designed to denoise the posterior distribution of the ground truth (GT) and predict GT conditions. The loss function of CRT is defined as follows:
\begin{equation}
    \mathcal{L}_{prior} = ||\boldsymbol{I}_{G} \odot \boldsymbol{m}_{f} - CRT(\boldsymbol{m}_{f} \odot \boldsymbol{I}'_{G_{t-1}}, \boldsymbol{y}_T, t)||^2_2 \label{eq:15}
\end{equation}
where $CRT(\cdot)$ stands for the CRT function, $\boldsymbol{I}_{G}$ represents the GT image and $\boldsymbol{I}'_{G_{t-1}}$ is obtained by Eq. \ref{eq:3}. Building upon the assumptions presented in~\cite{wang2023dr2}, it is assumed that there exists a time step $\gamma$ such that for $t > \gamma$, the distance between $q(\boldsymbol{x}_{t} |\boldsymbol{x})$ and $q(\boldsymbol{y}_{t} |\boldsymbol{y})$ becomes sufficiently small. At this point, we can get the approximate estimate:
\begin{equation}
\begin{aligned}    
    \boldsymbol{y}_{t-1} &= CRT(\boldsymbol{m}_{f} \odot \boldsymbol{y}'_{t-1}, \boldsymbol{y}_T, t)\\
    &\approx CRT(\boldsymbol{m}_{f} \odot \boldsymbol{I}'_{G_{t-1}}, \boldsymbol{y}_T, t)
    \label{eq:16}
\end{aligned}    
\end{equation}

\begin{figure}[t]
  \centering
  \includegraphics[width=0.99\linewidth]{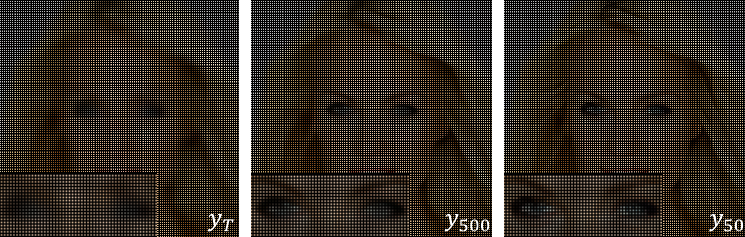}
  \caption{\textbf{Condition Refinement}. We present the initial condition $\boldsymbol{y}_{T}$ along with the refined conditions in the intermediate stages ($\boldsymbol{y}_{500}, \boldsymbol{y}_{50}$). Alongside the diffusion sampling, the conditions are updated with CRT. Refinement of the conditions contributes to a more effective sampling. Please zoom in for a better viewing.}
  \vspace{-0.45cm}
\label{fig3} 
\end{figure}

\begin{table}[t]
  \centering
  \setlength{\tabcolsep}{0.8mm}{
  \scalebox{0.58}{
      \begin{tabular}{c|c c c c|c c c c|c c c c}
        \toprule[1.5pt]
         & \multicolumn{4}{c|}{$\times4$} & \multicolumn{4}{c|}{$\times8$} & \multicolumn{4}{c}{$\times16$} \\
         Methods & LPIPS$\downarrow$ & FID$\downarrow$ & IS$\uparrow$ & PSNR$\uparrow$ & LPIPS$\downarrow$ & FID$\downarrow$ & IS$\uparrow$ & PSNR$\uparrow$ & LPIPS$\downarrow$ & FID$\downarrow$ & IS$\uparrow$ & PSNR$\uparrow$ \\
        \midrule[1.5pt]
        Dataset & \multicolumn{12}{c}{\textbf{CelebA1000}}\\
        \midrule[1.2pt]
         Bicubic    & 0.1257 & 67.94 & 2.25 & 28.97 & 0.2263 & 80.16 & 2.22 & 24.95 & 0.2989 & 158.38 & 2.38 & 21.65 \\
         CodeFormer  & \textcolor{blue}{0.0912} & \textcolor{blue}{17.03} & 2.73 & 27.45 & 0.1460 & 23.12 & 2.66 & 23.89 & 0.2162 & 34.15 & 2.50 & 20.75 \\
          DDNM & 0.1093 & 24.64 & 2.65 & \textcolor{blue}{29.00} & 0.1468 & 29.12 & 2.66 & 24.88 & 0.2074 & 32.38 & 2.42 & 21.56 \\
         DDRM & 0.1122 & 30.13 & 2.51 & 28.07 & 0.1589 & 36.34 & 2.34 & 24.81 & 0.2088 & 42.48 & 2.14 & 21.61 \\
         SR3 & 0.0970 & 33.25 & 2.62 & \textcolor{red}{29.27} & \textcolor{blue}{0.1435} & 56.69 & 2.45 & \textcolor{red}{25.48} & 0.2574 & 76.45 & 2.24 & \textcolor{red}{21.71} \\
         ILVR & 0.0931 & 21.18  & 2.41 & 27.74 & 0.1446 & 25.53 & 2.24 & 24.21 & \textcolor{blue}{0.2001} & 33.49 & 2.21 & 20.95 \\
         DR2 & 0.1767 & 52.58 & \textcolor{red}{3.07} & 23.23 & 0.1957 & 52.52 & \textcolor{red}{2.85} & 22.31 & 0.2243 & 53.18 & \textcolor{red}{2.79} & 20.57 \\
         DPS~ & 0.1497 & 22.04 & 2.36 & 23.96 & 0.1893 & 23.48 & 2.31 & 21.53 & 0.2303 & \textcolor{red}{25.83} & 2.34 & 19.20 \\
         DiffFace~ & 0.1048 & 19.86 & 2.65 & 27.68 & 0.1553 & 24.09 & 2.63 & 24.49 & 0.2167 & 31.80 & 2.52 & 20.32 \\
         PGDiff~ & 0.0918 & 18.78 & 2.86 & 28.96 & 0.1439 & \textcolor{red}{22.49} & 2.91 & 24.53 & 0.2003 & 27.68 & 2.55 & 21.24 \\
        \midrule
         \textbf{DPI(Ours)} & \textcolor{red}{0.0894} & \textcolor{red}{16.90} & \textcolor{blue}{2.89} & 28.07 & \textcolor{red}{0.1401} & \textcolor{blue}{22.86} & \textcolor{blue}{2.96} & \textcolor{blue}{24.97} & \textcolor{red}{0.1927} & \textcolor{blue}{27.33} & \textcolor{blue}{2.66} & \textcolor{blue}{21.68} \\
        \bottomrule[1.5pt]
      \end{tabular}
  }}
  \caption{Quantitative comparisons on \textbf{CelebA1000} testset. \textcolor{red}{\textbf{Red}} and \textcolor{blue}{blue} indicates the best and the second best.}\label{Tab:1} \vspace{-0.2cm}

\end{table}

However, the CRT is essentially a discriminative model that favors the average output and the gap between $q(\boldsymbol{x}_{t} |\boldsymbol{x})$ and $q(\boldsymbol{y}_{t} |\boldsymbol{y})$ progressively increases for $t < \gamma$. For this reason, we need to adapt the CRT to the intermediate conditions as well as mitigate the gap. Ultimately, the objective function for training the CRT is as follows:
\begin{align}
\mathcal{L}_{gap} &= ||\boldsymbol{I}_{G} \odot \boldsymbol{m}_{f} - CRT(\boldsymbol{m}_{f} \odot \boldsymbol{\hat x}_{crt}, \boldsymbol{y}_T, t)||^2_2 \label{eq:17}\\
\mathcal{L}_{crt} &=  \Omega \mathcal{L}_{prior} + (1 - \Omega) \mathcal{L}_{gap} \label{eq:18}
\end{align}
where $\boldsymbol{\hat x}_{crt}$ represents the intermediate condition output of CRT and $\Omega$ is a weight that decreases over $t$. The structure of CRT can be found in the Appendix 2. In the first stage, due to the sufficiently heavy input noise, CRT can only extract information from $\boldsymbol{y}_{T}$. The output of CRT during this stage is smooth. In the second stage, as the noise decreases and RACM amplifies the prior space, CRT can utilize prior information to update conditions.

\begin{figure*}[ht]
    \centering
    \includegraphics[height=0.20\textwidth]{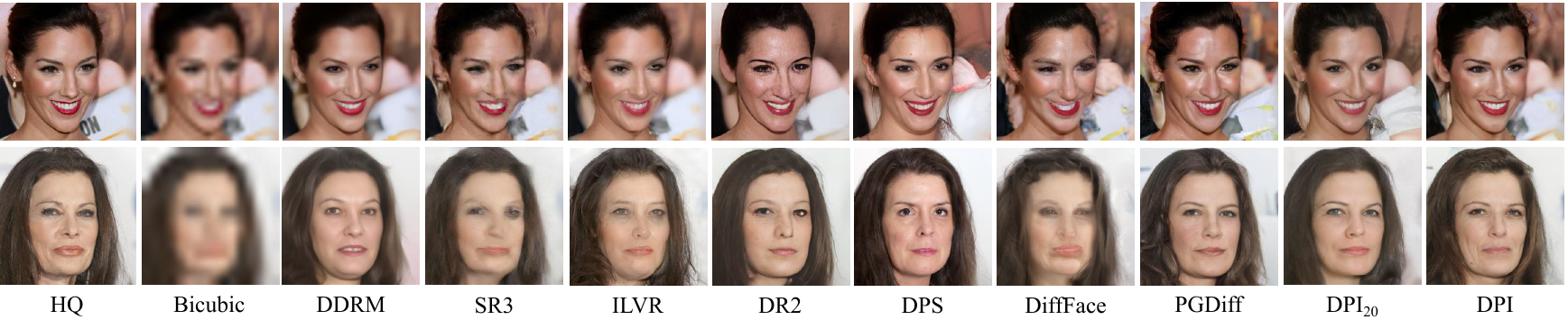}
  \caption{From top to bottom are the FSR results of DPI and DDPM-based methods on \textbf{CelebA1000} testset at $\times$8, and $\times$16 scales, respectively. $\text{DPI}_{20}$ refers that the DDIM~\cite{song2020denoising} algorithm samples only 20 steps. Please zoom in for best view.}
\label{fig4}  \vspace{-0.3cm} 
\end{figure*}

\begin{figure*}[t]
    \centering
    \includegraphics[height=0.23\textwidth]{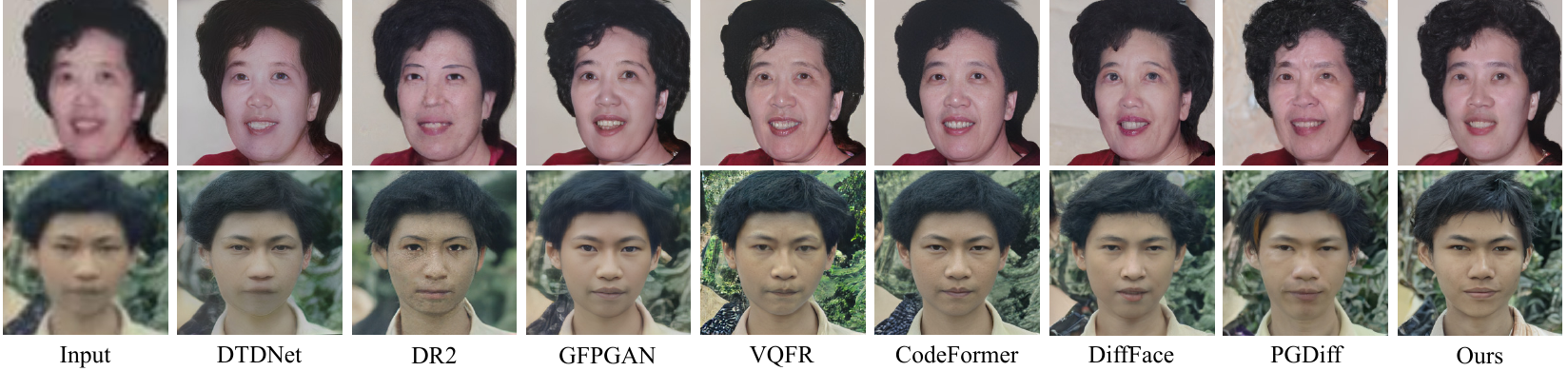}
  \caption{Qualitative comparisons on \textbf{WebPhoto} dataset. Please zoom in for best view.}
\label{fig5} \vspace{-0.3cm}
\end{figure*}

\begin{table}[t]
  \setlength{\tabcolsep}{0.8mm}{
        \scalebox{0.50}{\begin{tabular}{l|cccc|cccc|cccc}
        \toprule
         Datasets & \multicolumn{4}{c|}{LFW} & \multicolumn{4}{c|}{WebPhoto} & \multicolumn{4}{c}{WIDER} \\
         Degradation & \multicolumn{4}{c|}{\textit{mild}} & \multicolumn{4}{c|}{\textit{medium}} & \multicolumn{4}{c}{\textit{heavy}} \\       
         Methods & FID$\downarrow$ & IS$\uparrow$ & MUSIQ$\uparrow$ & CLIPIQA$\uparrow$ & FID$\downarrow$ & IS$\uparrow$ & MUSIQ$\uparrow$ & CLIPIQA$\uparrow$ & FID$\downarrow$ & IS$\uparrow$ &MUSIQ$\uparrow$ & CLIPIQA$\uparrow$ \\
        \midrule
         Input & 138.41 & 2.85 & 29.41 & 0.4432 & 163.24 & 3.28 & 21.04 & 0.3351 & 192.17 & 2.88 & 15.72 & 0.2976\\
         DFDNet & 87.83 & 3.52 & 65.92 & \textcolor{red}{0.7313} & 125.22 & 3.60 & 61.53 & 0.6526 & 135.78 & 3.27 & 56.78 & 0.6249 \\
         GFPGAN   & / & / & / & / & 133.73 & 3.45 & 63.78 & 0.6742 & 99.51 & \textcolor{blue}{3.48} & 63.07 & 0.6611 \\
         VQFR     & 69.02 & 3.53 & 64.44 & 0.6964 & 92.81 & 3.46 & 63.92 & 0.6933 & / & / & / & /\\
         CodeFormer & 68.72 & 3.56 & \textcolor{red}{67.30} & 0.7034 & 85.93 & 3.56 & \textcolor{red}{65.87} & \textcolor{red}{0.7050} & 51.61 & 3.34 & 64.48 & 0.7132 \\
         DR2 & 80.94 & 3.45 & 59.23 & 0.6095 & 99.27 & 3.53 & 54.53 & 0.5469 & 71.85 & 3.01 & 55.06 & 0.5739 \\
         DiffFace  & 66.40 & 3.55 & 63.57 & 0.6813 &  89.81 & 3.58 & 59.61 & 0.6624 & 50.36 & 3.41 & 58.56 & 0.6764\\
         PGDiff  & \textcolor{red}{62.31} & \textcolor{blue}{3.60} & 65.97 & \textcolor{blue}{0.7137} & \textcolor{blue}{85.83} & \textcolor{blue}{3.61} & 62.87 & 0.6950 & \textcolor{red}{47.63} & 3.47 & \textcolor{blue}{64.56} & \textcolor{blue}{0.7137} \\
        \midrule
         DPI & \textcolor{blue}{65.91} & \textcolor{red}{3.64} &  \textcolor{blue}{66.13} & 0.7098 & \textcolor{red}{81.77} & \textcolor{red}{3.67} & \textcolor{blue}{64.65} & \textcolor{blue}{0.7011} & \textcolor{blue}{49.79} & \textcolor{red}{3.50} & \textcolor{red}{64.90}& \textcolor{red}{0.7150}\\
         \midrule
      \end{tabular}}}
      \caption{Quantitative comparisons on the \textbf{real-world} datasets. \textcolor{red}{Red} and \textcolor{blue}{blue} indicate the best and the second best.}
    \label{tab2} \vspace{-0.5cm}
\end{table}

\begin{table}[t]
    \centering
        \scalebox{0.77}{\begin{tabular}{l|cc|cc|cc}
        \toprule
         & \multicolumn{2}{c|}{$\times4$} & \multicolumn{2}{c|}{$\times8$} & \multicolumn{2}{c}{$\times16$} \\
         Methods & CS$\downarrow$ & ACC$\uparrow$ & CS$\downarrow$ & ACC$\uparrow$ & CS$\downarrow$ & ACC$\uparrow$ \\
        \midrule
         Bicubic & 0.1061 & 90.2 & 0.3276 & 78.3 & 0.6475 & \textcolor{blue}{60.0}\\
         CodeFormer & 0.0991 & \textcolor{blue}{94.8} & 0.2253 & \textcolor{blue}{92.2} & \textcolor{blue}{0.3950} & 53.5\\
         DDRM   & 0.1466 & 92.8 & 0.2781 & 87.7 & 0.4155 & 47.5\\
         SR3     & \textcolor{blue}{0.0831} & 93.7 & 0.2264 & 91.2 & 0.4015 & 49.9\\
         ILVR    & 0.0898 & 91.5 & 0.2363 & 90.8 & 0.3998 & 53.3\\
         DR2     & 0.2455 & 89.4 & 0.3264 & 76.9 & 0.3955 & 52.8\\
         DPS     & 0.2470 & 92.2 & 0.3603 & 66.6 & 0.4550 & 31.8\\
         PGDiff  & 0.0870 & 93.1 & \textcolor{blue}{0.2103} & 90.8 & 0.4064 & 59.8\\
        \midrule
         DPI     & \textcolor{red}{0.0807} & \textcolor{red}{95.3} & \textcolor{red}{0.2100} & \textcolor{red}{92.5} & \textcolor{red}{0.3449} & \textcolor{red}{72.8}\\
         \midrule
      \end{tabular}}
    \caption{Qualitative comparisons of \textbf{face recognition} accuracy and consistency on CelebA1000 testset. \textcolor{red}{Red} and \textcolor{blue}{blue} indicate the best and the second best performance.}
    \label{tab3} \vspace{-0.5cm}
\end{table}

\section{Experiments}
\subsection{Experimental Setup}
We employ a pre-trained DDPM from DPS~\cite{chung2022diffusion}. The model is pre-trained on 49k face images from FFHQ~\cite{karras2019style} at a resolution of $256 \times 256$. For evaluation, we utilize synthetic datasets FFHQ1000 and CelebA1000~\cite{liu2015deep}, along with real-world datasets LFW~\cite{huang2008labeled}, WebPhoto~\cite{wang2021towards}, and WIDER~\cite{yang2016wider}, serving as our testsets. Following the settings of the latest diffusion-based methods~\cite{wang2023dr2,chung2022diffusion}, we compare them on synthetic datasets at scales of $\times4$, $\times8$, and $\times16$. Additionally, for each of these three scales, the parameters ($\tau$, $s$, $\omega$) is set to (100, 1.4, 500), (300, 1.2, 750), and (500, 1, 1000) respectively. For real-world datasets, we adhere to the experimental settings in CodeFormer~\cite{zhou2022towards}, with fixed parameters set to (500, 1, 1000). The sparsity parameter $k$ for CMs is set to 2 for all experiments. 

\begin{figure*}[!ht]
    \centering
    \includegraphics[height=0.132\textwidth]{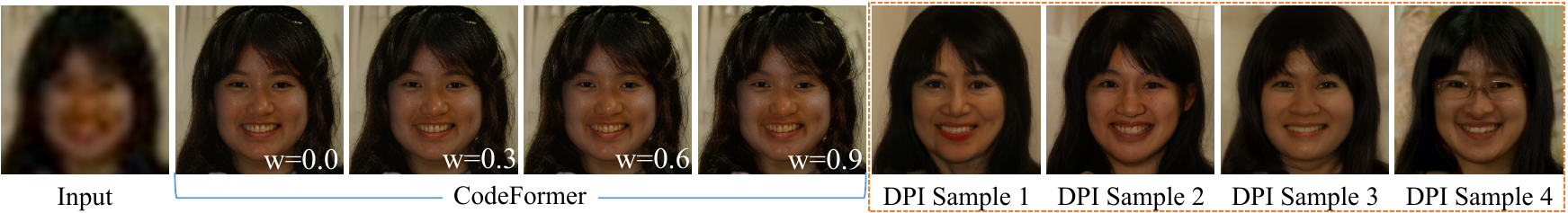}
  \caption{\textbf{Diversity Comparison in WIDER dataset}. DPI allows users to adjust hyperparameters to generate diverse results. Note that DPI can also be configured with hyperparameters to ensure consistency. Please zoom in for best view.}
\label{fig6} \vspace{-0.3cm}
\end{figure*}

\subsection{Comparison with Previous Work}
\textbf{Synthetic Datasets}: 
Diffusion-based FSR works typically use Bicubic settings of different scales as benchmarks. For this purpose, we train CRT at different scales to compare them~\cite{wang2022zero, kawar2022denoising, saharia2023image,choi2021ilvr,wang2023dr2, chung2022diffusion, kim2022diffface, yang2024pgdiff}. Table \ref{Tab:1} demonstrates the outstanding performance of DPI across all scales, showcasing SOTA perceptual metrics compared to other diffusion-based methods. Fig. \ref{fig4} illustrates DPI's superior performance in visual consistency. \\ 
\textbf{Real-world Datasets}: We adhere to the general degradation model~\cite{zhou2022towards} represented as follows: 
\begin{equation}
    \boldsymbol{I}_L=\{[(\boldsymbol{I}_H \otimes k_{s,\sigma})_{\downarrow r}+n_{\delta}]_{\text{JEPG}_q}\}_{\uparrow r}
    \label{eq19}
\end{equation}
where $k_{s, \sigma}$ denotes a Gaussian blur kernel with a kernel size of $s$, $n_{\delta}$ represents Gaussian noise, $\text{JPEG}_q$	signifies compression with quality $q$, and $r$ denotes the sampling scale. CRT is trained on this degradation model to address real-world issues. Our DPI achieves SOTA performance in no-reference metrics on three real-world datasets, as shown in Table \ref{tab2}. We demonstrate the diversity generation capability of DPI in Fig. \ref{fig6}. \\
\textbf{Face Recognition Results}: We employ the open-source face recognition framework, DeepFace~\cite{serengil2020lightface} with a threshold set at 0.4 to compare the accuracy (ACC) and consistency (CS) of face recognition between GT and SR images. Specifically, we feed the SR results and GT images into the DeepFace model to determine if they correspond to the same person. ACC represents the proportion of accurate recognition in the testset, while CS represents the distance between SR and GT features. Table \ref{tab3} presents the experimental results on the CelebA1000 testset for different upscaling factors. Especially in the $\times$16 task, we are far ahead of other methods. The results of face recognition proves that DPI has higher consistency.  \\
\textbf{Severe Degradation Blind FSR}: For severe blind FSR, Eq. \ref{eq19} is applied to sample parameters $r, s, \sigma, q, \delta$ from $\{8 : 16\}$, $\{1 : 17\}$, $\{3 : 20\}$, $\{40 : 50\}$, $\{30 : 90\}$ to construct the testset. The hyperparameters of DPI are consistent with real-world tasks. As illustrated in Fig. \ref{fig7}, it can be seen that DPI can ensure the consistency of the contour even under severe degradation, which reflects the strong robustness. \\

\begin{figure}[t]
    \centering
    \includegraphics[width=1\linewidth]{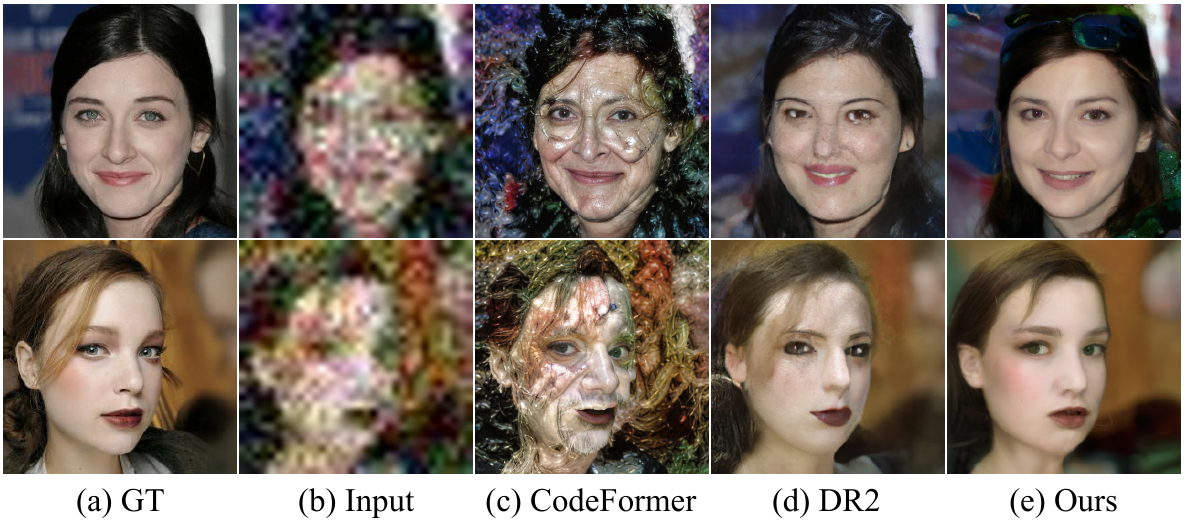}
  \caption{Qualitative comparison on heavy degradation.}
  \label{fig7} \vspace{-0.5cm}
\end{figure}

\begin{table}[t]
    \centering
\scalebox{0.8}{\begin{tabular}{lcccccc} \\\toprule  
w/ CRT  & \checkmark &  & \checkmark & \checkmark & \checkmark & \checkmark\\
w/ Refine & \checkmark & \checkmark &  & \checkmark & \checkmark & \checkmark\\
w/ Stage II & \checkmark & \checkmark & \checkmark &  & \checkmark & \checkmark\\
w/ Weight & \checkmark & \checkmark & \checkmark &  &  & \checkmark\\
w/ $k=2$ & \checkmark & \checkmark & \checkmark & \checkmark & \checkmark & \\
\midrule
SSIM$\uparrow$ & 0.6881 & 0.6933 & 0.6873 & \textbf{0.7278} & 0.6362 & 0.6335\\
LPIPS$\downarrow$ & \textbf{0.1401} & 0.2242 & 0.1507 & 0.1486 & 0.1679 & 0.1592\\
FID$\downarrow$  & \textbf{22.86} & 73.01 & 23.65 & 53.60 & 33.78 & 28.09\\\bottomrule
\end{tabular}}
\vspace{-3mm}
\captionof{table}{Ablation studies}
\label{Tab:4} \vspace{-0.5cm}
\end{table}

\begin{figure}[t]
    \centering
  \includegraphics[width=1\linewidth]{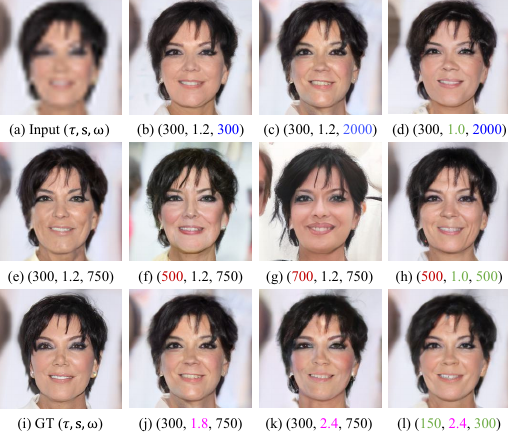}
  \caption{Visual comparisons of the impact of hyperparameters on consistency and diversity.}
  \label{fig8} \vspace{-0.5cm}
\end{figure}

\subsection{Ablation Studies}

In this section, we conduct ablation experiments to analyze the effects of strong and weak constraints, masking strategies, and hyperparameters on consistency and diversity. All ablation studies are performed on the CelebA1000 testset at a scale of $\times$8. Please refer to the visual ablation studies in Appendix D for further insight. \\ 
\textbf{Effectiveness of Condition Correction and Refinement}: The conditions undergo pixel correction and iterative refinement to provide accurate guidance. Initially, the degraded pixels' impact on performance is analyzed by not using the CRT to repair the conditions. Subsequently, the corrected condition are used solely as conditions at each time step without participating in the iterative refinement. As shown in the the second and third columns of Table \ref{Tab:4}, degraded pixels lead to poor metrics and significantly affect image quality, while the absence of refinement only marginally impacts performance. \vspace{1mm} \\
\textbf{Balancing Consistency and Diversity}: Our proposed method allows the user to tune the consistency and diversity of FSR by simply adjusting the scalar values, as shown in Fig. \ref{fig8}. In our default settings (Fig. \ref{fig8} (e)), the parameters $(\tau, s, \omega)$ are set to (300, 1.2, 750). We will elaborate on the impact of these scalars on the results: 1) By adjusting $\tau$, we can flexibly partition the sampling stages. Increasing the range of the second stage significantly enhances diversity, while conversely improving consistency. As seen in the fourth column of Table \ref{Tab:4}, removing the second stage ($\tau = 0$) initially yields the best SSIM, but the FID metric noticeably increases, indicating enhanced consistency but reduced fidelity. The second row of Fig. \ref{fig8} (e-g) clearly demonstrates that increasing $\tau$ leads to greater diversity. 2) Under the premise of ensuring consistency (Fig. \ref{fig8} (e, j, k)), we fine-tune the diversity by changing the sparsity. 3) In Eq. \ref{eq:9}, $\omega$ is used to reduce conditional intensity for better utilization of prior information. A larger $\omega$ implies a smaller weight attached to the condition. From the comparison in Fig. \ref{fig8} (b) and (c), reducing conditions intensity enhances fidelity and diversity, such as details in hair. The ablation study in the sixth column of Table \ref{Tab:4} also confirms this. Furthermore, the comparison of Fig. \ref{fig8} (f) and Fig. \ref{fig8} (h) reflects the control of $s$ and $\omega$ over consistency and diversity.

\section{Conclusion}

We propose DPI, which effectively leverages the prior knowledge of pre-trained models for face super-resolution. By implementing a masking strategy tailored to facial features, we achieve a balance between consistency and diversity during the sampling process. Additionally, we introduce CRT to establish a reciprocal sampling process, where samples and conditions are iteratively refined. Extensive experiments demonstrate the superior performance of DPI and its ability to ensure consistency.

\section{Acknowledgement}
This work is supported in part by the National Natural Science Foundation of China, under Grant (62302309, 62171248), Shenzhen Science and Technology Program (JCYJ20220818101014030, JCYJ20220818101012025).

\bibliography{aaai25}

\clearpage
\appendix

\section{Appendix}
In the appendix, we provide additional experimental details, quantitative experiments, and analyses. We begin by introducing the seamless integration of DPI into the DDIM algorithm. Subsequently, we offer a more comprehensive description of our experiments to highlight the fairness and rigor. We also include additional quantitative experiments and complexity analyses. Finally, we present visual results from ablation studies and discuss the processing of natural images.

\section{A. Diffusion Prior Interpolation with DDIM}

Algorithm \ref{alg2} demonstrates the utilization of DDIM~\cite{song2020denoising} for our DPI. As the sampling interval is compressed, it necessitates the adjustment of parameters. We set the hyperparameters ($\tau$, $s$, $w$) for different tasks ($\times$4, $\times$8, $\times$16, \text{real}) as (6, 1.8, 15), (7, 1.4, 30), (7, 1.4, 60) and (7, 1.4, 60) respectively. $\eta$ is set to 0.1 for all experiments. It is noteworthy that the condition Corrector (CRT) has not undergone retraining with respect to the rescaled noise scheduler. As shown in Fig. \ref{fig2}, we present visualization results at different steps. It can be observed that with an increase in the number of steps, the images exhibit higher fidelity.

\begin{algorithm}[hbp]
\caption{Diffusion Prior Interpolation with DDIM, given an uncontional diffusion model $\epsilon_{\theta}(\boldsymbol{x}_t, t)$, Corrector $CRT(\cdot)$ and number of implicit sampling iterations $T$.}
\label{alg2}
\textbf{Input}: $\boldsymbol{y}_T$, $\tau$, $s$, $\omega$, $\eta$, $\boldsymbol{m}_f$ \\
$\boldsymbol{x}_T \gets$ sample from $\mathcal{N}(\mathbf{0}, \mathbf{I})$
\begin{algorithmic}[1] 
\FORALL{$t$ from $T$ to 1}
\STATE $\sigma_t^2 = \eta \sqrt{\frac{(1-\bar{\alpha}_{t-1})}{1-\bar{\alpha}_t}}\sqrt{\frac{1-\bar{\alpha}_t}{\bar{\alpha}_{t-1}}}$
\STATE $\boldsymbol{z}_t \gets$ sample from $\mathcal{N}(\mathbf{0}, \mathbf{I})$
\STATE $\boldsymbol{x}'_{t-1} = \sqrt{\bar{\alpha}_{t-1}}\left(\frac{\boldsymbol{x}_t-\sqrt{1-\bar{\alpha}_t}\boldsymbol{\epsilon}_\theta(\boldsymbol{x}_t, t)}{\sqrt{\bar{\alpha}_t}}\right) + \sqrt{1-\bar{\alpha}_{t-1}-\sigma^2_t}\boldsymbol{\epsilon}_\theta(\boldsymbol{x}_t, t) +\sigma_tz_t$
\STATE $\boldsymbol{y}'_{t-1} = \sqrt{\bar{\alpha}_{t-1}}\left(\frac{\boldsymbol{y}_t-\sqrt{1-\bar{\alpha}_t}\boldsymbol{\epsilon}_\theta(\boldsymbol{x}_t, t)}{\sqrt{\bar{\alpha}_t}}\right) + \sqrt{1-\bar{\alpha}_{t-1}-\sigma^2_t}\boldsymbol{\epsilon}_\theta(\boldsymbol{x}_t, t) +\sigma_tz_t$
\IF{$t > \tau$}
\STATE $\boldsymbol{x}_{t-1} = (1 - \boldsymbol{m}_f) \odot \boldsymbol{x}'_{t-1} + \boldsymbol{m}_f \odot \boldsymbol{y}'_{t-1}$
\ELSE
\STATE $\boldsymbol{m}_a = Mask_{ga}(\boldsymbol{y}_t, s)$
\STATE $w = t / \omega$
\STATE 
$\boldsymbol{x}_{t-1} = (1 - \boldsymbol{m}_a) \odot \boldsymbol{x}'_{t-1} + w  \boldsymbol{m}_a \odot \boldsymbol{y}'_{t-1} +$ \\
\quad \quad \quad$ (1-w) \boldsymbol{m}_{a} \odot \boldsymbol{x}'_{t-1}$
\STATE $\boldsymbol{y}'_{t-1} = \boldsymbol{x}_{t-1}$
\ENDIF
\STATE $\boldsymbol{y}_{t-1} = CRT(\boldsymbol{m}_f \odot \boldsymbol{y}'_{t-1}, \boldsymbol{m}_f \odot \boldsymbol{y}_T, t)$
\ENDFOR
\RETURN $\boldsymbol{x}_0$
\end{algorithmic}
\end{algorithm}

\begin{figure}[t]
  \small
\centering
  \includegraphics[width=1\linewidth]{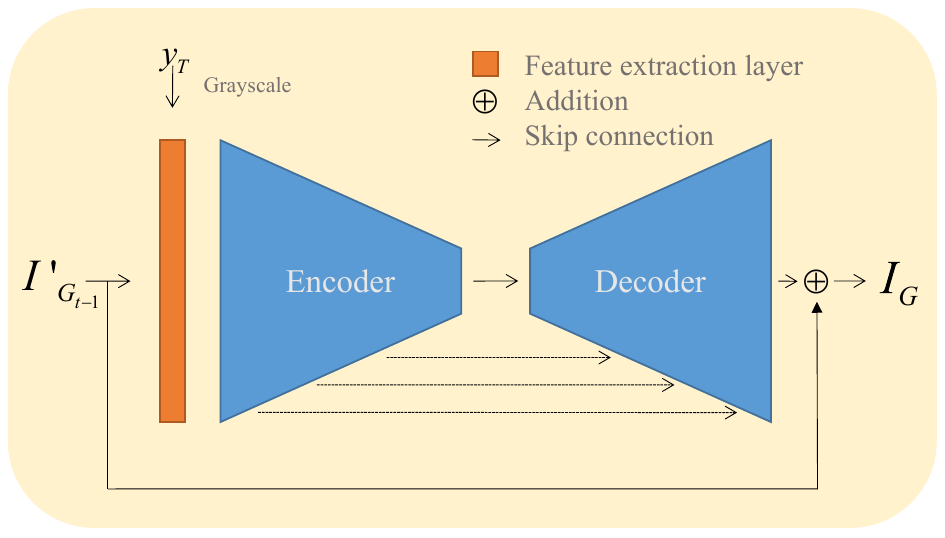}
  \caption{\textbf{Illustrative architecture of Corrector.}}
  \label{fig1}
\end{figure}

\begin{figure*}[!h]
    \centering
    \includegraphics[height=0.6\textwidth]{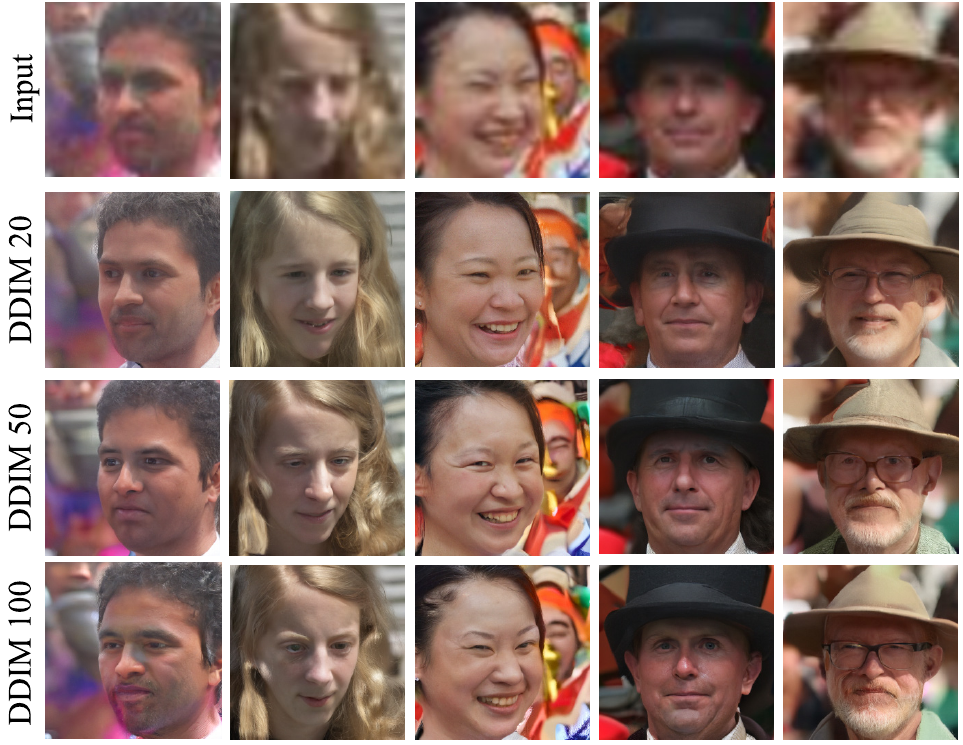}
  \caption{Quantitative comparison at different steps on \textbf{WIDER} dataset.}
\label{fig2} 
\end{figure*}

\begin{table*}[t]
\begin{center}
\begin{small}
\captionof{table}{Complexity analysis of different methods}
\scalebox{0.99}{
\begin{tabular}{cccccccc}
            \toprule
            & ILVR & DDRM & $\text{DDNM}^{+}$ & DR2 & DPS & PGDiff & Ours \\
            \midrule
             Blind or Real-world? & No & No & Partial & Yes & No & Yes & Yes \\
             NFEs & 1000 & 20 & 100 & 30 & 1000 & 2000 & 20 \\
             Single-step Time & \textcolor{red}{32ms}  & +2ms & +25ms & +57ms & +36ms & +39ms & +18ms \\
             Memory & 1196MB & 1146MB & 1554MB & 2794MB & 2896MB & 4796MB & 1354MB \\
             Additional Parameters & 0 & 0 & 0 & 85.79M & 0 & 16.77M & 5.86M \\
            \bottomrule
          \end{tabular}}
        \label{tab:com}
\end{small}
\end{center}
\end{table*}

\begin{table}[t]
    \centering
    \scalebox{0.72}{\begin{tabular}{l|cc|cc|cc}
        \toprule
         & \multicolumn{2}{c|}{$\times4$} & \multicolumn{2}{c|}{$\times8$} & \multicolumn{2}{c}{$\times16$} \\
         Methods & CS$\downarrow$ & ACC$\uparrow$ & CS$\downarrow$ & ACC$\uparrow$ & CS$\downarrow$ & ACC$\uparrow$ \\
        \midrule
         Bicubic & 0.1295 & 85.1 & 0.3226 & 74.2 & 0.5844 & 11.3 \\
        CodeFormer & 0.0806 & 93.2 & 0.1935 & 91.9 & 0.3109 & 80.5\\
         DDRM  & 0.1420 & 88.8 & 0.2383 & 84.6 & 0.3414 & 72.5\\
         SR3  & \textcolor{red}{0.0627} & \textcolor{blue}{94.2} & \textcolor{blue}{0.1797} & \textcolor{blue}{92.0} & 0.3222 & 70.1 \\
         ILVR & 0.0944 & 89.5 & 0.1884 & 88.8 & 0.3013 & 80.4\\
         DR2  & 0.1872 & 93.3 & 0.2388 & 89.8 & \textcolor{blue}{0.2942} & \textcolor{blue}{81.6}\\
         DPS  & 0.1970 & 90.4 & 0.2744 & 85.7 & 0.3704 & 62.2\\
         PGDiff  & 0.0823 & 93.8 & 0.1938 & 91.2 & 0.3003 & 77.5\\
        \midrule
         DPI   & \textcolor{blue}{0.0781} & \textcolor{red}{96.1} & \textcolor{red}{0.1790} & \textcolor{red}{92.3} & \textcolor{red}{0.2733} & \textcolor{red}{84.8}\\
         \midrule
      \end{tabular}}
    \caption{Qualitative comparisons of \textbf{face recognition} accuracy and consistency on FFHQ1000 testset. \textcolor{red}{Red} and \textcolor{blue}{blue} indicate the best and the second best performance.}
    \label{tab1}
\end{table}

\section{B. Condition Corrector}
The condition Corrector (CRT) proposed in this paper employs a U-Net architecture sourced from~\cite{dhariwal2021diffusion}. We utilize the Adam~\cite{kingma2014adam} optimizer with parameters $\beta_1$ = 0.9 and $\beta_2$ = 0.999 for training. Throughout all experiments, we employ an exponential moving average (EMA) decay rate of 0.9999. The PyTorch~\cite{paszke2019pytorch} framework is employed, and the training is conducted in parallel on 3080Ti GPUs with a batch size of 32. For all experiments, CRT is trained on the FFHQ dataset using various degradation models. For instance, in synthetic experiments, we train CRT at different scales, while for real-world experiments, we train CRT based on the degradation model described in Eq. 19. CRT is applicable to any pre-trained face diffusion model. Furthermore, the model architecture of CRT is not constrained to a specific structure.

We train the network using the GT posterior distribution $\boldsymbol{I}_{G_{t-1}}$ as input, with the initial condition $\boldsymbol{y}_T$ and time $t$ as conditions, as depicted in Fig. \ref{fig1}. The grayscale $\boldsymbol{y}_T$ is directly added to the feature extraction layer, and we apply residual learning. This approach offers the advantage that the CRT consistently produces accurate conditions, regardless of the noise intensity in the posterior distribution. Specifically, the CRT can extract meaningful features from the input or condition $\boldsymbol{y}_T$ to produce the desired output. Moreover, conventional techniques that combine discriminative and generative models, such as~\cite{whang2022deblurring,shang2023resdiff}, typically involve generating initial blurry images using MSE-based estimations, followed by enriching details using generative models to attain high-quality images. However, these methods lack the capability to refine samples alongside pre-trained models. Our proposed method adheres to the paradigm of diffusion model sampling and seamlessly integrates into the posterior, enabling the CRT to refine the conditions throughout the entire sampling process, providing precise guidance.

\begin{table*}[t]
  \centering
    \resizebox{\textwidth}{30mm}{
      \begin{tabular}{c|c c c c|c c c c|c c c c}
        \toprule[1.5pt]
         & \multicolumn{4}{c|}{$\times4$} & \multicolumn{4}{c|}{$\times8$} & \multicolumn{4}{c}{$\times16$} \\
         Methods & LPIPS$\downarrow$ & FID$\downarrow$ & PSNR$\uparrow$ & SSIM$\uparrow$ & LPIPS$\downarrow$ & FID$\downarrow$ & PSNR$\uparrow$ & SSIM$\uparrow$ & LPIPS$\downarrow$ & FID$\downarrow$ & PSNR$\uparrow$ & SSIM$\uparrow$ \\
         \midrule[1.5pt]
         Dataset & \multicolumn{12}{c}{\textbf{FFHQ-1000}}\\
         \midrule[1.2pt]
         Bicubic & 0.1562 & 129.91 & \textcolor{blue}{28.27} & \textcolor{red}{0.8211} & 0.2529 & 126.76 & 24.43 & 0.6919 & 0.3284 & 224.25 & 21.18 & 0.5818 \\
         CodeFormer  & \textcolor{blue}{0.0979} & 48.67 & 26.70 & 0.7742 & 0.1552 & 57.79 & 23.36 & 0.6467 & 0.2239 & 69.42 & 20.46 & 0.5314 \\
          DDNM  & 0.1175 & 42.21 & 25.56 & 0.7677 & 0.1686 & 55.64 & 24.37 & 0.6853 & 0.2139 & 49.64 & 21.67 & 0.5771 \\
         DDRM  & 0.1242 & 50.49 & 27.59 & 0.7924 & 0.1743 & 61.21 & 24.29 & \textcolor{red}{0.7003} & 0.2260 & 73.08 & 21.11 & \textcolor{red}{0.6079} \\
         SR3 & 0.1037 & 40.04 & \textcolor{red}{28.69} & 0.8060 & 0.1575 & 70.98 & \textcolor{red}{24.86} & 0.6782 & 0.2445 & 83.16 & \textcolor{red}{21.39} & 0.5121 \\
         ILVR & 0.1007 & \textcolor{blue}{31.08}  & 27.19 & \textcolor{blue}{0.8112} & 0.1543 & 35.85 & 23.74 & 0.6747 & 0.2096 & 40.95 & 20.49 & 0.5774 \\
         DR2 & 0.1785 & 47.30 & 22.66 & 0.6184 & 0.2004 & 48.11 & 21.77 & 0.5833 & 0.2279 & 46.52 & 20.09 & 0.5253 \\
         DPS & 0.1583 & 35.77 & 23.38 & 0.6732 & 0.1978 & 35.40 & 20.97 & 0.5864 & 0.2414 & \textcolor{red}{36.09} & 18.66 & 0.5034 \\
         DiffFace & 0.1061 & 37.77 & 26.36 & 0.7336 & 0.1689 & 39.74 & 23.88 & 0.6479 & 0.2226 & 40.64 & 19.93 & 0.5355 \\
          PGDiff & 0.0991 & 33.12 & 27.79 & 0.7688 & \textcolor{blue}{0.1543} & \textcolor{blue}{33.29} & 23.99 & 0.6797 & \textcolor{blue}{0.2088} & 38.36 & 20.96 & 0.5654 \\
         
        \midrule
         \textbf{DPI(Ours)} & \textcolor{red}{0.0965} & \textcolor{red}{30.16} & 27.32 & 0.7742 & \textcolor{red}{0.1505} & \textcolor{red}{31.22} & \textcolor{blue}{24.54} & \textcolor{blue}{0.6853} & \textcolor{red}{0.2081} & \textcolor{blue}{37.07} & \textcolor{blue}{21.27} & \textcolor{blue}{0.5847} \\
        \bottomrule[1.5pt]
      \end{tabular}
      }
  \caption{Quantitative comparisons on \textbf{FFHQ1000} dataset. \textcolor{red}{Red} and \textcolor{blue}{blue} indicates the best and the second best performance.}
  \label{tab2}
\end{table*}

\begin{figure*}[t]
    \centering
    \includegraphics[height=0.3\textwidth]{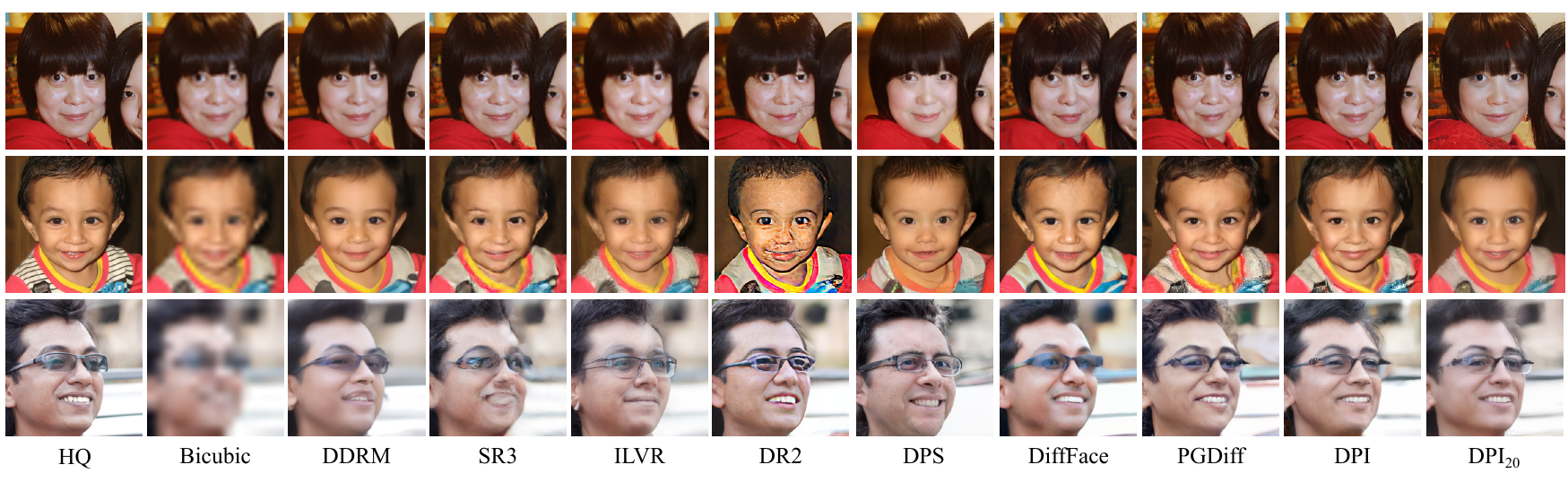}
  \caption{FSR visualizations of DPI and other diffusion-based methods on \textbf{FFHQ1000} testset. The results of FSR under $\times$4, $\times$8, and $\times$16 downsampling from top to bottom. Please zoom in for best view.}
\label{fig3} 
\end{figure*}

\begin{figure*}[t]
    \centering
    \includegraphics[height=0.45\textwidth]{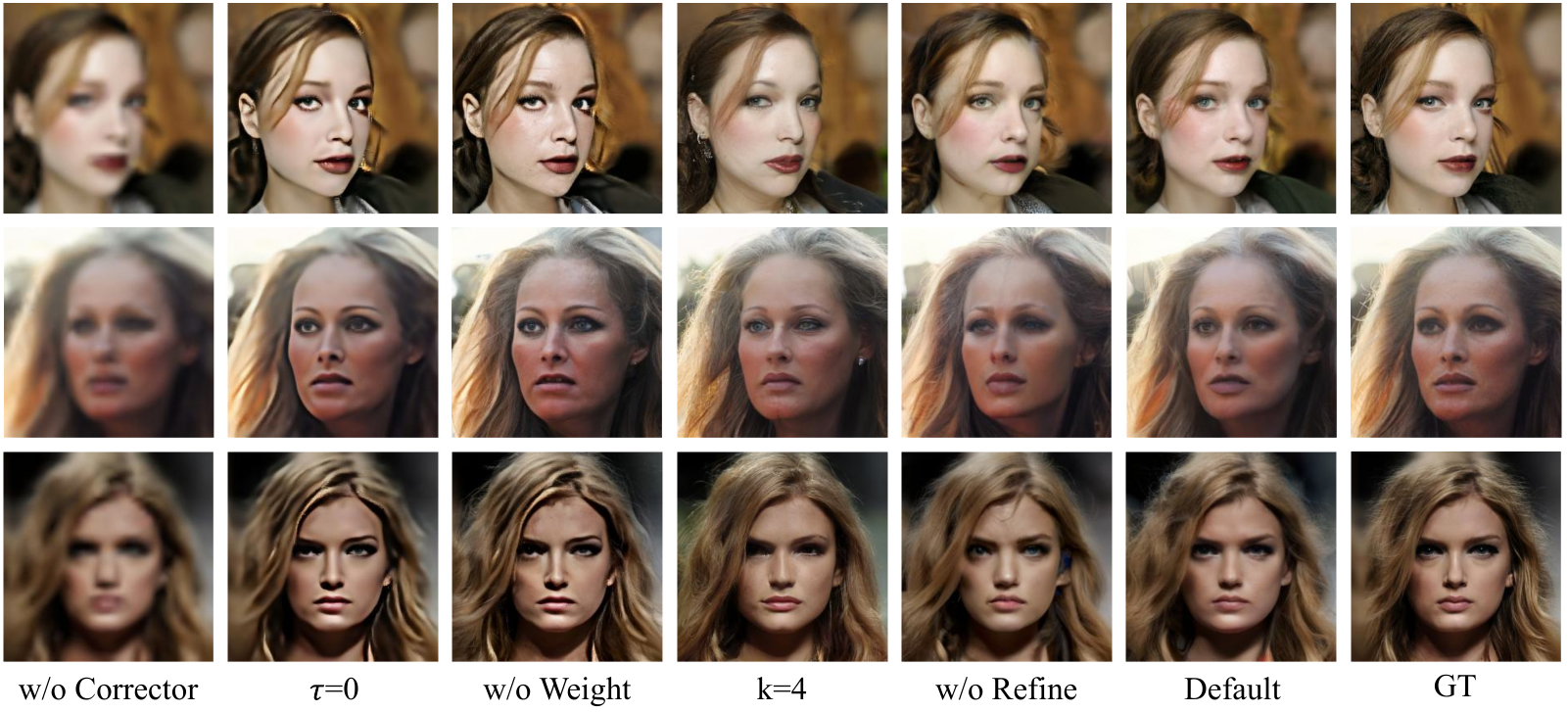}
  \caption{\textbf{Visualizations of Ablation Studies.}}
\label{fig6} \end{figure*}

\begin{figure*}[t]
    \centering
    \includegraphics[height=0.315\textwidth]{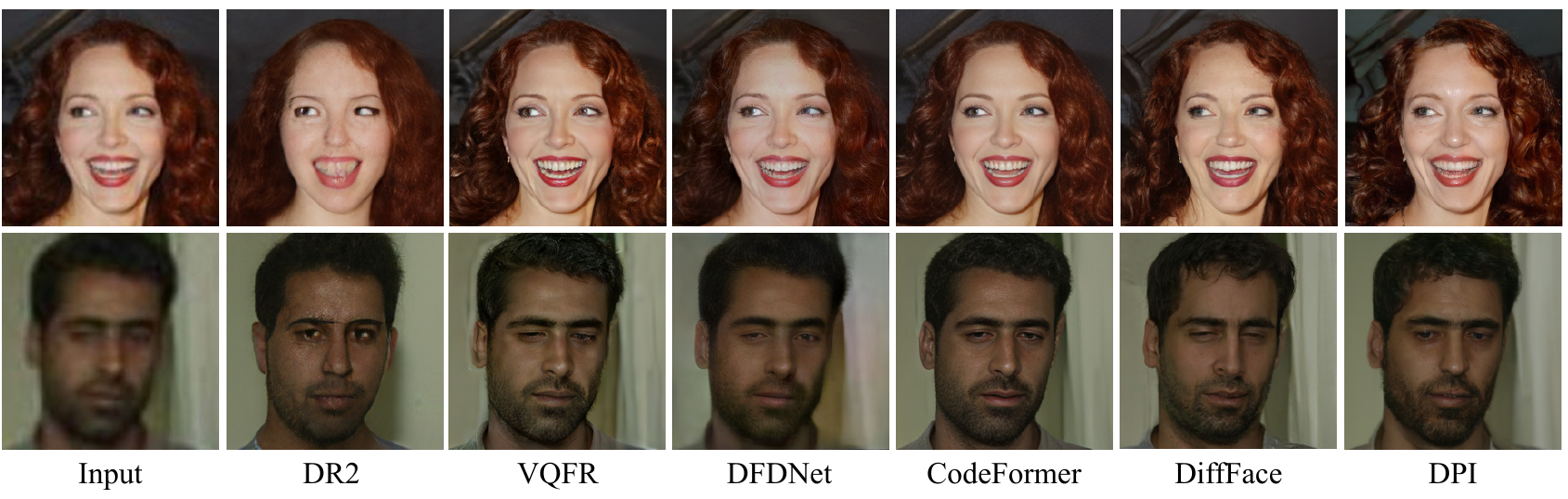}
  \caption{Qualitative comparisons on \textbf{LFW} dataset.}
  \label{fig4}
\end{figure*}

\begin{figure*}[t]
    \centering
    \includegraphics[height=0.37\textwidth]{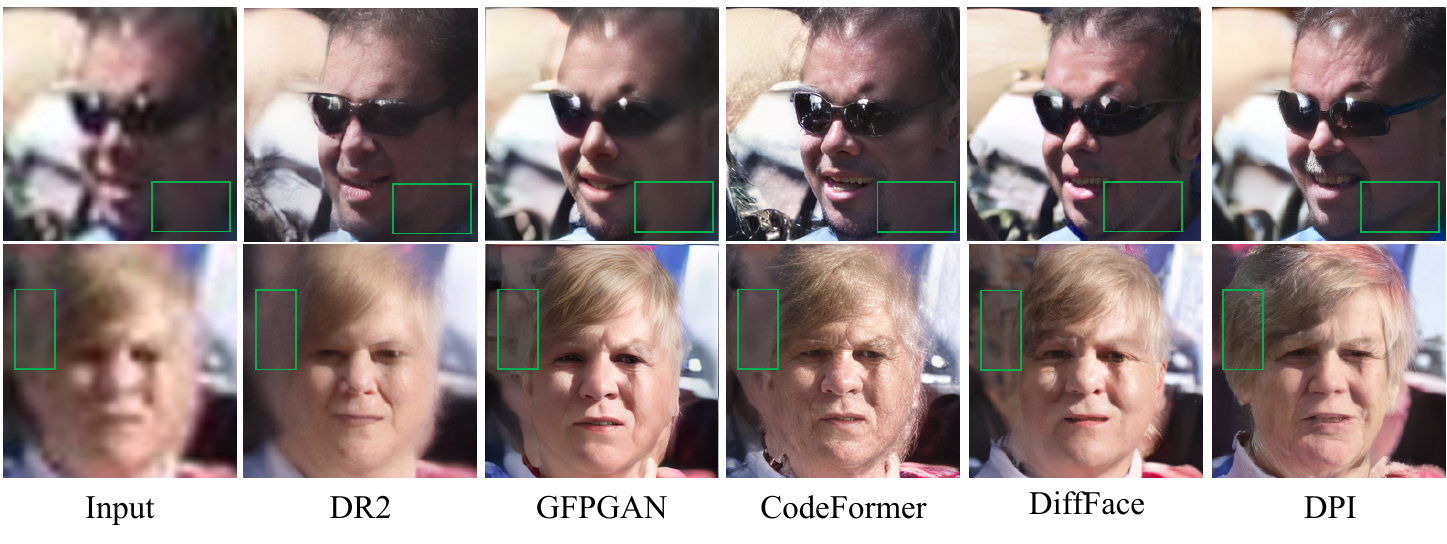}
  \caption{Qualitative comparisons on \textbf{WIDER} dataset.}
  \label{fig5}
\end{figure*}

\begin{figure*}[t]
    \centering
    \includegraphics[height=0.34\textwidth]{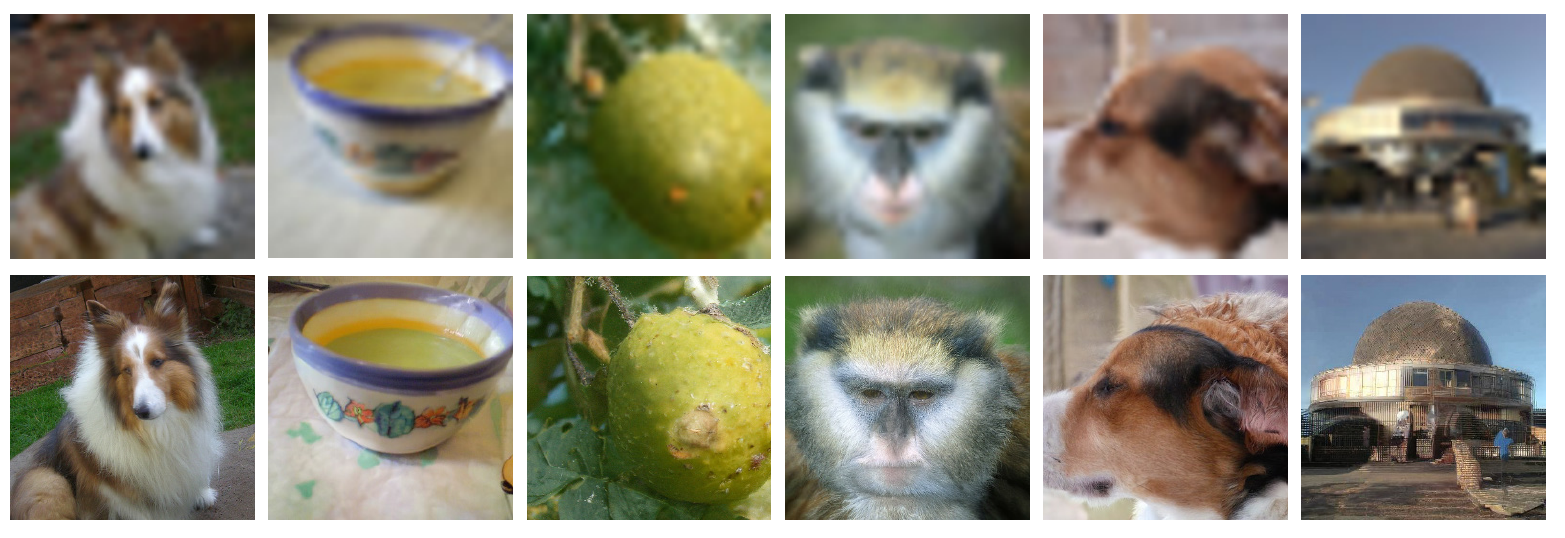}
  \caption{Visualization of general image super-resolution. The top and bottom images correspond to the LR and SR images, respectively.}
  \label{fig7}
\end{figure*}

\section{C. More Experiments}
\label{more}

We conduct comprehensive experiments to validate the superiority of our algorithm compared to other diffusion-based methods, including DDRM~\cite{kawar2022denoising}, DDNM~\cite{wang2022zero}, SR3~\cite{saharia2023image}, ILVR~\cite{choi2021ilvr}, DR2~\cite{wang2023dr2}, DPS~\cite{chung2022diffusion}, DiffFace~\cite{kim2022diffface}, and PGdiff~\cite{yang2024pgdiff}. Among these approaches, only SR3 requires training from scratch, while the others are based on pre-trained models. Our method, along with DDRM and other diffusion-based methods, utilizes the same pre-trained model and model weights from DPS. The training code for SR3 is derived from a standard implementation\footnote{\url{https://github.com/Janspiry/Image-Super-Resolution-via-Iterative-Refinement}}, which is also employed in other works~\cite{chung2022diffusion,yang2024pgdiff}. The training data for SR3 is consistent with the pre-trained model, specifically FFHQ 49k. Previous diffusion-based works have not addressed real-world face super-resolution. We compare our approach with CNN/Transformer-based methods, including DFDNet~\cite{li2020blind}, GFPGAN~\cite{wang2021towards}, VQFR~\cite{gu2022vqfr}, and CodeFormer~\cite{zhou2022towards}.

Table \ref{tab1} displays the performance comparison of different methods for face recognition accuracy and face consistency on the FFHQ1000 testset. The visual results and the quantitative evaluations on the FFHQ1000 testset are presented in Fig. \ref{fig3} and Table \ref{tab2}, respectively. DPI achieves comparable performance in comparison with the SOTA method. We also present comparative results on real datasets LFW and WIDER, as shown in Fig. \ref{fig4} and Fig. \ref{fig5}. In the experiment with heavy degradation (Fig. \ref{fig5}), it can be observed that other methods are prone to introducing errors and artifacts into the FSR results. Our DPI, due to the randomness of RACM, can mitigate this issue to some extent.

\section{D. Complexity analysis}

Since SR3 and DiffFace require training a diffusion model from scratch, we do not compare their complexity with other methods. Most other works rely on the Guided Diffusion framework and pre-trained weights\footnote{\url{https://github.com/openai/guided-diffusion}}, which allows for a fair complexity analysis. Methods based on pre-trained diffusion models generally have similar generation speeds to the original models. The differences in their complexity primarily arise from the following aspects:

\begin{itemize}
    \item Conditioning Mechanism: The introduction of conditioning adds extra computational and memory overhead during sampling steps. The complexity introduced by sophisticated conditioning methods becomes more pronounced as iterations accumulate.
    \item Additional Neural Networks: Utilizing additional neural networks to guide the diffusion prior enables the handling of more complex tasks. The time complexity of these networks directly impacts the overall complexity of the method.
    \item Accelerated Sampling: The primary time complexity in diffusion-based methods stems from the Number of Function Evaluations (NFEs). When there is no additional forward diffusion, NFEs are equivalent to the number of sampling steps.
\end{itemize}

We conduct a comprehensive complexity analysis, as shown in Table 1. These methods can be categorized into two groups based on their ability to handle blind or real-world tasks. Although $\text{DDNM}^{+}$ addresses some real-world scenarios, its performance is suboptimal, and it lacks quantitative metrics on real-world datasets. ILVR introduces conditioning with constant complexity, resulting in inference time closest to the original DDPM. Using ILVR as a baseline, we compare the single-step time complexity increases caused by the introduction of conditioning in different methods. Our method, like DR2 and PGDiff, incurs additional time overhead due to the inclusion of extra neural networks. However, compared to DPS, which requires gradient computation, and DDNM+, which involves inner loops, our method is competitive in terms of speed. Additionally, our CRT network model is relatively small, resulting in highly efficient parameter and memory usage.

\section{E. Visualizations of Ablation Studies}
Here, we present visualizations of ablation studies to better understand the impact of hyperparameters on the balance between consistency and diversity. From left to right in Fig. \ref{fig6}, several key observations can be made. Firstly, it is evident that the application of the CRT has the most pronounced impact on image quality (first column). This phenomenon arises from the fact that the degradation cues impose erroneous guidance, causing the results to conform closely to the manifold of the LR representation. When $\tau=0$, we furnish the entire sampling with the corrected Fix Conditional Mask (FCM). While the FCM offers commendable consistency guidance, it is worth noting that deficiencies in MSE estimator, compounded by the high intensity of FCM, contribute to generating suboptimal results in regions demanding intricate textural details (second column). The results of applying the Randomly Adaptive Conditional Mask (RACM) without weighted are demonstrated in Fig. \ref{fig6}, appearing in the third column, wherein heightened texture details in high-frequency regions such as hair and eyes are evident.

Nevertheless, it remains evident that these SR images lack fine-grained verisimilitude grounded in real priors. To harness the real priors effectively, we manipulate the sparsity of the RACM to widen the utilization of priors. With $k$ set to 4, we observe that the SR images exhibit more realistic texture details. Alternatively, we can incorporate real priors by reducing the weight of the RACM, as indicated by Eq. 9. The results of the default setting showcase the outcome of reduced-weight RACM. In contrast to the comparison with increased sparsity of the RACM, increased sparsity promotes greater diversity, while weight reduction strikes a better balance between consistency and diversity. 

Lastly, we emphasize the significance of the condition refinement. The comparisons in the fifth and sixth columns reveal that the absence of refinement has a minor impact on results while maintaining good consistency and fidelity similar to the default setting. However, the absence of refinement may induce color deviations due to the lack of updating the condition in line with real priors.

\section{F. Discussion and Limitations}

The real-world degradation model encompasses various downstream tasks such as denoising, deblurring, image enhancement, and super-resolution. Our approach is effective in handling real-world and mixed degradation scenarios, meaning that DPI is capable of addressing individual sub-tasks efficiently. Moreover, DPI is akin to prior-based inpainting methods~\cite{song2020score}. Consequently, we do not focus on the specific sub-task of face restoration. In this work, we design a conditional mask specifically for facial priors and validate it across numerous face benchmarks. We find that the distribution of natural images is highly diverse and rich in high-frequency information. Without relying on supervised fine-tuning, it is challenging to ensure high consistency of results solely through conditional prior constraints. For instance, the prior-based diffusion work cited in this paper still requires further exploration for natural image restoration. Furthermore, methods that utilize supervised fine-tuning, such as StableSR~\cite{wang2024exploiting}, still face challenges in achieving consistency.

DPI can be seamlessly adapted for natural image SR. We train a CRT on ImageNet using the degradation model based on Eq. 19 to handle natural images. The hyperparameters in DPI and CRT are consistent with those used for the real-world FSR. We validate natural images using a DDPM pre-trained on ImageNet from DPS~\cite{chung2022diffusion}, as shown in Fig. \ref{fig7}. Since facial features are relatively fixed and lack complex background information, we can effectively use masking strategies to ensure consistency. However, the details and backgrounds of natural images are more complex. As a result, the effectiveness of ensuring consistency through RACM constraints is diminished.

\end{document}